\documentclass[sigconf]{acmart}

\usepackage{array}
\usepackage{url}
\usepackage{bm}
\usepackage{rotating}
\usepackage{booktabs}
\usepackage{amsfonts}
\usepackage{amsmath,amssymb}
\newcommand{\st}{\mathrm{s.t.}}

\newcommand{\tr}{\mathrm{tr}}

\newcommand{\tabincell}[2]{\begin{tabular}{@{}#1@{}}#2\end{tabular}}
\usepackage{multirow}
\usepackage{algorithm}
\usepackage{algorithmic}

\usepackage{graphicx,subfig}
\usepackage{float}

\AtBeginDocument{%
  \providecommand\BibTeX{{%
    \normalfont B\kern-0.5em{\scshape i\kern-0.25em b}\kern-0.8em\TeX}}}

\setcopyright{acmcopyright}
\copyrightyear{2020}
\acmYear{2020}
\acmDOI{xx.xxxx/xxxxxxx.xxxxxxx}

\acmConference[ACAI '20]{International Conference on Algorithms, Computing and Artificial Intelligence}{December 24--26, 2020}{Sanya, China}
\acmBooktitle{ACAI '20: International Conference on Algorithms, Computing and Artificial Intelligence,
  December 24--26, 2020, Sanya, China}
\acmPrice{15.00}
\acmISBN{978-1-4503-8811-5}

\begin{document}

\title{Manifold Adaptive Multiple Kernel K-Means for Clustering}

\author{Liang Du}
\affiliation{%
	\department{School of Computer and Information Technology}
	\institution{Shanxi University}
	\city{Taiyuan Shi}
	\state{Shanxi Sheng}
	\country{China}}
\email{duliang@sxu.edu.cn}
\author{Haiying Zhang}
\affiliation{%
	\department{School of Computer and Information Technology}
	\institution{Shanxi University}
	\city{Taiyuan Shi}
	\state{Shanxi Sheng}
	\country{China}}
\email{1021528684@qq.com}

\author{Xin Ren}
\affiliation{%
	\department{School of Computer and Information Technology}
	\institution{Shanxi University}
	\city{Taiyuan Shi}
	\state{Shanxi Sheng}
	\country{China}}
\email{710372070@qq.com}
\author{Xiaolin Lv}
\affiliation{%
	\department{School of Computer and Information Technology}
	\institution{Shanxi University}
	\city{Taiyuan Shi}
	\state{Shanxi Sheng}
	\country{China}}
\email{969103861@qq.com}

\renewcommand{\shortauthors}{Du and Zhang, et al.}

\begin{abstract}
Multiple kernel methods based on k-means aims to integrate a group of kernels to improve the performance of kernel k-means clustering. However, we observe that most existing multiple kernel k-means methods exploit the nonlinear relationship within kernels, whereas the local manifold structure among multiple kernel space is not sufficiently considered. In this paper, we adopt the manifold adaptive kernel, instead of the original kernel, to integrate the local manifold structure of kernels. Thus, the induced multiple manifold adaptive kernels not only reflect the nonlinear relationship but also the local manifold structure. We then perform multiple kernel clustering within the multiple kernel k-means clustering framework. It has been verified that the proposed method outperforms several state-of-the-art baseline methods on a variety of data sets.
\end{abstract}



\terms{Algorithms}


\maketitle

\section{Introduction}
Clustering is one of the fundamental topics in data mining, machine learning and pattern recognition. Instead of conducting data clustering within the original feature space, kernel clustering methods perform clustering within the Reproducing Kernel Hilbert space (RKHS), where the nonlinear relationship may be better captured. One of the newly introduced problem for kernel clustering is the design or selection of the proper kernel function, where the best choice is data dependent and unknown in advance.

The Multiple Kernel Clustering (MKC) methods are with great potential to alleviate the effort for kernel designing or integrating complementary information \cite{yu2012optimized} by leveraging a predefined set of candidates kernels from different functions or views. It is natural to extend existing single kernel clustering methods into multiple kernel scenario. The typical methods include K-means based \cite{huang2012multiple,yu2012optimized,lmkc_nips,rmkkm_ijcai2015,mkkmmr_2016,onmkkm_2017,zhou2019incremental,lmkkm_2018}, self-organizing map (SOM) \cite{mkc_som}, maximum margin clustering based \cite{mmc_nips2004,mmc_nips2006,mkc_sdm2009}, local learning-based  \cite{zeng2011feature}, spectral clustering based \cite{ctsc,coreg_nips2011,huang2012affinity, DBLP:conf/ijcai/FanCZWD17,cka_pr2014, mksc_icdm2012,rmsc_2014} and subspace clustering based \cite{twsc_2017,kdsl_2017,uscog_2018,rmkcss_2020,mkc_nkss_2019} algorithms. Compared with the single kernel counterpart, MKC should take special effort to handle the additional data problems such as noisy and incomplete kernels \cite{rmkkm_ijcai2015,rmsc_2015,wang2015experimental,mkcck_2017,lgrmkc_2018,rmkcss_2020,lmkkm_2018,pami_2019_01,pami_2019_02}.

Although the candidate kernel well capture the similarities among samples in different nonlinear feature spaces, it dose not necessary characterize the underlying local geometric structure of data, which is vital important for unsupervised learning tasks.	Moreover, only a few efforts \cite{lkamkc_2016,glsamkc,mkc_nkss_2019} have been taken to incorporate the local geometric structure of data for MKC. It is worthwhile to point out that these methods use the combined consensus kernel matrix to compute the local graphs, where the discrete neighborhood relationships are changed during the optimization procedure. 

In this paper, we present the novel Manifold Adaptive Multiple Kernel K-Means for data clustering (MAMKKC). Given the input kernel matrix, we first construct the corresponding nearest neighbor graph to capture the underlying manifold structure. Then we incorporate the manifold structure into the kernel space via the manifold adaptive kernel mechanism. As a result, the newly induced kernel matrices not only reflect the nonlinear relationship but also the manifold structure. Finally, we linearly combine these manifold adaptive kernels within the multiple kernel k-means clustering framework. We also derive the corresponding optimization procedure to reduce the objective function monotonically and obtain the optimal solution for the proposed MAMKKC model. It has been verified that the proposed method outperforms several state-of-the-art baseline methods on a variety of data sets.	


\section{Manifold Adaptive Multiple Kernel K-Means}
 In this paper, we define the number of samples, clusters, kernels as $n, c, m$, respectively. Suppose that in the clustering task, let $\{\mathcal{K}^{p}\}_{p=1}^{m}$ denotes the $m$ different kernel functions. correspondingly, there must be $m$ different associated feature spaces denoted as $\{\mathcal{H}^{p}\}_{p=1}^m$. The purpose of multiple kernel clustering is to generate the final clustering result via the integration of multiple candidate kernels.

In this paper, we adopt the manifold adaptive kernel transformation \cite{DBLP:conf/icml/SindhwaniNB05} to embed the local manifold structure within kernels. Let $\mathcal{V}$ be a linear space with a positive semi-definite inner product (quadratic form) and let $\mathcal{S}$:$\mathcal{H} \to \mathcal{V}$ be a bounded linear operator. We define  $\widetilde{\mathcal{H}}$ to be the space of functions from $\mathcal{H}$ with the modified inner product
\begin{align}
	\langle f,g \rangle_{\widetilde{\mathcal{H}}} =\langle f,g \rangle_\mathcal{H}+\langle Sf,Sg \rangle_\mathcal{H}
\end{align}
It has been shown that $\widetilde{\mathcal{H}}$ is still a RKHS.

Given $f=(f(x_1),\dots,f(x_n))^T$ and $g=(g(x_1),\dots,g(x_n))^T$. Notice that $f,g \in \mathcal{V}$,thus we have
\begin{align}
	\langle Sf,Sg \rangle_{\mathcal{V}}=\langle \mathbf{f},\mathbf{g} \rangle=f^T \mathbf{M} g.
\end{align}
where $\mathbf{M}$ is a positive semi-definite matrix, and 
\begin{align}
	\mathbf{k}_{\mathbf{x}}=\left(\mathcal{K}(\mathbf{x},\mathbf{x}_1),\cdots,\mathcal{K}(\mathbf{x},\mathbf{x}_n)\right).
\end{align}
It can be shown that the reproducing kernel kernel in $\widetilde{\mathcal{H}}$ is
\begin{align}
	\widetilde{\mathcal{K}}(\mathbf{x},\mathbf{z})=\mathcal{H}(\mathbf{x},\mathbf{z})-\lambda {\mathbf{k}^T_\mathbf{x}} (\mathbf{I}+\mathbf{M}\mathbf{K})^{-1}\mathbf{M}\mathbf{k}_\mathbf{z}	
\end{align}
where $\mathbf{I}$ is an identity matrix, $\mathbf{K}$ is the kernel matrix in $\mathcal{H}$ and $\lambda \geqslant 0$ is a constant controlling the smoothness of the functions. In this paper, we construct a series of $\tau$ nearest neighbor graphs $\{\mathbf{G}^p\}_{p=1}^{m}$ to reflect the manifold structure for each candidate kernel. Then, we can construct the associated graph Laplacian  $\{\mathbf{L}^p\}_{p=1}^{m}$ via  $\mathbf{L}^{p}=\mathbf{D}^{p}-\mathbf{G}^{p}$. Let $\mathcal{K}$ be any data-independent kernel associated with the kernel matrix $\mathbf{K}$. That is, $\mathbf{K}_{ij}=\mathcal{K}(\mathbf{x}_i,\mathbf{x}_j)$. Let $\mathbf{k}_i$ be the $i-$th column vector of $\mathbf{K}$. 	
By setting $\mathbf{M}=\mathbf{L}$, we can calculate the manifold adaptive kernel matrix $\mathcal{K}_{\mathcal{M}}$ as follows
\begin{align}
	\mathcal{K}_{\mathcal{M},ij}=\mathbf{K}_{ij}-\lambda {\mathbf{k}_i}^T (\mathbf{I}+\mathbf{L}\mathbf{K})^{-1}\mathbf{L}\mathbf{k}_j.	
\end{align}
It is important to note that all the candidate kernels can be transformed to manifold adaptive kernels. In the setting of multiple kernel clustering, we can get many different manifold adaptive kernels, and we linearly integrate these deformed manifold adaptive kernels with different weight.
\begin{align}\label{compute_Km}
	\mathcal{K}_{\mathcal{M}}=\sum_{p=1}^{m} {w^2_p} \mathcal{K}^p_{\mathcal{M}}
\end{align}
where $\{w_p\}_{p=1}^{m}$ is the non-negative kernel weight.

Based on the kernel k-means clustering, we present the novel Manifold Adaptive Multiple Kernel K-Means for data clustering (MAMKKC), which can be formulated as follows
\begin{align}\label{opt_amkkc}
	\min_{\mathbf{Y}, \mathbf{w}} \quad & \tr\big(\sum_{p=1}^{m} {w_p}^2 \mathcal{K}^p_{\mathcal{M}} (\mathbf{I}-\mathbf{Y}\mathbf{Y}^T) \big) \\
	\st \quad & \mathbf{w} \geq 0, \sum_{p=1}^{m} w_p = 1,\mathbf{Y}^T \mathbf{Y}= \mathbf{I},
	\nonumber\\
	& \mathcal{K}_{\mathcal{M},ij}^{p}=\mathbf{K}_{ij}-\lambda {\mathbf{k}^T_i} \nonumber (\mathbf{I}+\mathbf{L}\mathbf{K})^{-1}\mathbf{L}\mathbf{k}_j,\\
	&\mathbf{L}=\mathbf{D}-\mathbf{W}.\nonumber
\end{align}
where $\tr$ is the trace function, $\lambda$ is the regularization parameter, $\mathbf{Y}$ is the scaled partition matrix, $\mathbf{w}$ is the weight of kernels, $\mathbf{I}$ is the identity matrix.		

\section{Optimization}
As seen from Eq.~\eqref{opt_amkkc}, there are two different variables need to be optimized. We adopted an alternate algorithm which is optimizing one variable while keeping the other fixed.

\subsection{Update $\mathbf{Y}$}
When the $\mathbf{w}$ is fixed, the optimization problem with regard to the variable $\mathbf{Y}$ can be formulated as follows
\begin{align}\label{opt_H}
	\max_{\mathbf{Y} } \quad &  \tr(\mathbf{Y}^T \mathbf{N} \mathbf{Y} )\\
	\st \quad & \mathbf{Y}^T \mathbf{Y}= \mathbf{I}, \nonumber
\end{align}
where $\mathbf{N}=\sum_{p=1}^{m}  {w}_p^2 \mathbf{K}^p$. The optimal $\mathbf{Y}^*$ fo the above trace maximization problem can be obtained by the eigenvectors corresponding to the $c$ largest eigenvalues of $\mathbf{N}$. The final discrete clustering result then can be obtained via k-means algorithm or spectral rotation from $\mathbf{Y}^*$.

\subsection{Update $\mathbf{w}$}
When the variable $\mathbf{Y}$ is fixed, the rest optimization problem with regard to the variable $\mathbf{w}$ can be formulated as:
\begin{align}\label{opt_w}
	\min_{w} \quad& \mathbf{w}^T \mathbf{A} \mathbf{w} \\
	\st \quad & \mathbf{w} \geq 0, \sum_{p=1}^{m} w_p = 1, \nonumber
\end{align}
where $\mathbf{A}$ is a diagonal matrix with diagonal element of $\mathbf{A}_{pp} = \tr(\mathbf{K}^p-\mathbf{K}^p \mathbf{Y} \mathbf{Y}^T)$. The above problem can be solved by off-the-shelf packages.

\subsection{Summarization of Algorithm}
In sum, we propose the iterative updating algorithm of optimizing Eq. \eqref{opt_amkkc} in Algorithm \ref{algo_onmkcf}.
\begin{algorithm}[ht]
	\caption{The algorithm to solve Eq.~\eqref{opt_amkkc}}
	\begin{algorithmic}[1]\label{algo_onmkcf}
		\REQUIRE {$\{ \mathbf{K}^{p} \}_{p=1}^{m}$,$\mathbf{Y}$, $\mathbf{w}$, $\lambda$,$\tau$}
		\REPEAT
		\STATE {Update $\mathbf{K}_\mathcal{M}$ according to Eq.~\eqref{compute_Km};}
		\STATE {Update $\mathbf{w}$ by solving Eq.~\eqref{opt_w};}
		\STATE {Update $\mathbf{Y}$ by solving Eq.~\eqref{opt_H};}
		\UNTIL{Converges}
		\ENSURE {$\mathbf{Y}$, $\mathbf{w}$}
	\end{algorithmic}
\end{algorithm}			
It is obvious that the problem in Eq.~\eqref{opt_amkkc} is lower bounded. The optimization with respect to $\mathbf{Y},\mathbf{w}$ will reduce the objective function in Eq.~\eqref{opt_amkkc} monotonically. Therefore, a local optimal solution can be expected according to our Algorithm \ref{algo_onmkcf}.

\subsection{Algorithm Complexity Analysis}
In this subsection, we will discuss the complexity of our proposed algorithm. The computation cost of computing $\tau$-nearest neighbors of all sample points in all the base kernels, i.e., $\{ \mathbf{G}^p\}_{p=1}^{m}$, is $\mathcal{O}(mn \tau^3)$. The computation cost of computing Laplacian matrices, i.e., $\{ \mathbf{L}^p\}_{p=1}^{m}$, is $\mathcal{O}(mn^2)$. The computation cost of computing $\{\mathbf{K}_\mathcal{M}^p\}_{p=1}^{m}$ is $\mathcal{O}(mn^3)$. The computation cost of one iteration for Eq.~\eqref{compute_Km}, Eq.~\eqref{opt_H} and Eq.~\eqref{opt_w} is $\mathcal{O}(mn^2 + m^3 + n^3)$. Suppose the total number of iteration is $t$, the overall computational cost for MAMKKC is $\mathcal{O}(mn \tau^3 + mn^2 + mn^3 + (mn^2 + m^3 + n^3)t)$. In our experiments, our algorithm converges very fast and the times $t$ is less than 20. Since $t \ll n$, $m \ll n$ and $\tau \ll n$, the total computational cost can be simplified as $\mathcal{O}(n^3)$.

\begin{table*}
	\caption{Clustering results measured by Accuracy/NMI/Purity of the compared methods}
	\centering \label{table:res-aio} \setlength{\tabcolsep}{2.0pt}
	\begin{tabular}{c c  c  c  c  c  c  c  c  c  c  c }
		\toprule
		Data Sets & Metrics & CTSC & Coreg & RMSC & RMKKM & MKKMMR & LKAMKC & ONMKC & LKGr & JMKSC & MAMKKC\\ \hline
		
		BBC& \tabincell{c}{ ACC\\NMI\\Purity}& \tabincell{c}{ 0.5577 \\  0.3783 \\  0.6024 }& \tabincell{c}{ 0.5414 \\  0.3664 \\  0.5943 }& \tabincell{c}{ 0.5373 \\  0.3699 \\  0.6404 }& \tabincell{c}{ 0.8019 \\  0.5963 \\  0.8019 }& \tabincell{c}{ 0.7436 \\  0.4736 \\  0.7436 }& \tabincell{c}{ 0.7476 \\  0.5216 \\  0.7476 }& \tabincell{c}{ 0.7463 \\  0.5051 \\  0.7463 }& \tabincell{c}{ 0.5142 \\  0.3778 \\  0.6106 }& \tabincell{c}{ 0.5617 \\  0.3310 \\  0.5984 }& \tabincell{c}{ \textbf{0.8114} \\ \textbf{0.6301} \\ \textbf{0.8114 }} \\ \hline
		RELATHE& \tabincell{c}{ ACC\\NMI\\Purity}& \tabincell{c}{ 0.6412 \\  0.0990 \\  0.6412 }& \tabincell{c}{ 0.6741 \\  0.1272 \\  0.6741 }& \tabincell{c}{ 0.5809 \\  0.0548 \\  0.5809 }& \tabincell{c}{ 0.5690 \\  0.0094 \\  0.5690 }& \tabincell{c}{ 0.6300 \\  0.0908 \\  0.6300 }& \tabincell{c}{ 0.7169 \\  0.1537 \\  0.7169 }& \tabincell{c}{ 0.6370 \\  0.0910 \\  0.6370 }& \tabincell{c}{ 0.5830 \\  0.0379 \\  0.5830 }& \tabincell{c}{ 0.5718 \\  0.0295 \\  0.5718 }& \tabincell{c}{ \textbf{0.8535} \\ \textbf{0.4169} \\ \textbf{0.8535 }} \\ \hline
		PIE10P& \tabincell{c}{ ACC\\NMI\\Purity}& \tabincell{c}{ 0.5286 \\  0.5973 \\  0.5333 }& \tabincell{c}{ 0.4333 \\  0.5349 \\  0.4810 }& \tabincell{c}{ 0.4143 \\  0.4444 \\  0.4333 }& \tabincell{c}{ 0.3143 \\  0.3596 \\  0.3286 }& \tabincell{c}{ 0.5095 \\  0.5799 \\  0.5333 }& \tabincell{c}{ 0.5981 \\  0.6561 \\  0.6314 }& \tabincell{c}{ 0.5905 \\  0.6061 \\  0.5952 }& \tabincell{c}{ 0.4381 \\  0.5101 \\  0.4429 }& \tabincell{c}{ 0.6333 \\  0.6908 \\  0.6381 }& \tabincell{c}{ \textbf{0.7524} \\ \textbf{0.8566} \\ \textbf{0.7952 }} \\ \hline
		COIL20& \tabincell{c}{ ACC\\NMI\\Purity}& \tabincell{c}{ 0.6917 \\  0.7762 \\  0.6931 }& \tabincell{c}{ 0.6764 \\  0.7737 \\  0.6917 }& \tabincell{c}{ 0.6806 \\  0.7848 \\  0.7076 }& \tabincell{c}{ 0.6736 \\  0.7646 \\  0.6958 }& \tabincell{c}{ 0.6868 \\  0.7850 \\  0.7188 }& \tabincell{c}{ 0.7007 \\  0.7843 \\  0.7194 }& \tabincell{c}{ 0.6937 \\  0.7961 \\  0.7181 }& \tabincell{c}{ 0.6868 \\  0.7634 \\  0.7028 }& \tabincell{c}{ 0.7479 \\  0.8540 \\  0.7826 }& \tabincell{c}{ \textbf{0.8153} \\ \textbf{0.9011} \\ \textbf{0.8535 }} \\ \hline
		UMIST& \tabincell{c}{ ACC\\NMI\\Purity}& \tabincell{c}{ 0.4852 \\  0.6545 \\  0.5426 }& \tabincell{c}{ 0.5026 \\  0.6835 \\  0.5530 }& \tabincell{c}{ 0.4817 \\  0.6813 \\  0.5426 }& \tabincell{c}{ 0.4504 \\  0.6538 \\  0.5339 }& \tabincell{c}{ 0.5026 \\  0.6874 \\  0.5774 }& \tabincell{c}{ 0.4974 \\  0.7011 \\  0.5861 }& \tabincell{c}{ 0.5391 \\  0.7250 \\  0.6209 }& \tabincell{c}{ 0.4887 \\  0.6915 \\  0.5687 }& \tabincell{c}{ 0.6104 \\  0.7575 \\  0.6748 }& \tabincell{c}{ \textbf{0.6609} \\ \textbf{0.8221} \\ \textbf{0.7530 }} \\ \hline
		BASEHOCK& \tabincell{c}{ ACC\\NMI\\Purity}& \tabincell{c}{ 0.9358 \\  0.6560 \\  0.9358 }& \tabincell{c}{ 0.9473 \\  0.7056 \\  0.9473 }& \tabincell{c}{ 0.9413 \\  0.6806 \\  0.9413 }& \tabincell{c}{ 0.9704 \\  0.8160 \\  0.9704 }& \tabincell{c}{ 0.9308 \\  0.6369 \\  0.9308 }& \tabincell{c}{ 0.9538 \\  0.7324 \\  0.9538 }& \tabincell{c}{ 0.9674 \\  0.7993 \\  0.9674 }& \tabincell{c}{ 0.6608 \\  0.0773 \\  0.6608 }& \tabincell{c}{ 0.5228 \\  0.0172 \\  0.5228 }& \tabincell{c}{ \textbf{0.9739} \\ \textbf{0.8273} \\ \textbf{0.9739 }} \\ \hline
		Leukemia& \tabincell{c}{ ACC\\NMI\\Purity}& \tabincell{c}{ 0.6806 \\  0.0970 \\  0.6806 }& \tabincell{c}{ 0.6111 \\  0.0508 \\  0.6528 }& \tabincell{c}{ 0.6250 \\  0.0586 \\  0.6528 }& \tabincell{c}{ 0.7083 \\  0.1216 \\  0.7083 }& \tabincell{c}{ 0.6389 \\  0.0670 \\  0.6528 }& \tabincell{c}{ 0.7222 \\  0.1148 \\  0.7222 }& \tabincell{c}{ 0.7361 \\  0.1611 \\  0.7361 }& \tabincell{c}{ 0.7083 \\  0.1216 \\  0.7083 }& \tabincell{c}{ 0.7222 \\  0.2602 \\  0.7222 }& \tabincell{c}{ \textbf{0.8194} \\ \textbf{0.2694} \\ \textbf{0.8194 }} \\ \hline
		ALLAML& \tabincell{c}{ ACC\\NMI\\Purity}& \tabincell{c}{ 0.7361 \\  0.1509 \\  0.7361 }& \tabincell{c}{ 0.6667 \\  0.0862 \\  0.6667 }& \tabincell{c}{ 0.5694 \\  0.0385 \\  0.6528 }& \tabincell{c}{ 0.7361 \\  0.1509 \\  0.7361 }& \tabincell{c}{ 0.6667 \\  0.1472 \\  0.6667 }& \tabincell{c}{ 0.7083 \\  0.2090 \\  0.7083 }& \tabincell{c}{ 0.7222 \\  0.1461 \\  0.7222 }& \tabincell{c}{ 0.7639 \\  0.1863 \\  0.7639 }& \tabincell{c}{ 0.7639 \\  0.1900 \\  0.7639 }& \tabincell{c}{ \textbf{0.8472} \\ \textbf{0.3416} \\ \textbf{0.8472 }} \\ \hline
		Average& \tabincell{c}{ mACC\\mNMI\\mPurity }& \tabincell{c}{ 0.6571 \\  0.4261 \\  0.6706 }& \tabincell{c}{ 0.6316 \\  0.4160 \\  0.6576 }& \tabincell{c}{ 0.6038 \\  0.3891 \\  0.6440 }& \tabincell{c}{ 0.6530 \\  0.4340 \\  0.6680 }& \tabincell{c}{ 0.6636 \\  0.4335 \\  0.6817 }& \tabincell{c}{ 0.7056 \\  0.4841 \\  0.7232 }& \tabincell{c}{ 0.7040 \\  0.4787 \\  0.7179 }& \tabincell{c}{ 0.6055 \\  0.3457 \\  0.6301 }& \tabincell{c}{ 0.6418 \\  0.3913 \\  0.6593 }& \tabincell{c}{ \textbf{0.8168} \\ \textbf{0.6332} \\ \textbf{0.8384 }} \\ \hline
		
		\bottomrule
	\end{tabular}
\end{table*}

\begin{table*}
	\caption{Clustering comparison on (mean ACC)/(standard derivation)/($p$-value). The bolder results are significant better than others ($p \le 0.05$).}
	\centering \label{table:acc-aio}\setlength{\tabcolsep}{1.0pt}
	\begin{tabular}{c c  c  c  c  c  c  c  c  c  c }
		\toprule
		Data Sets & CTSC & Coreg & RMSC & RMKKM & MKKMMR & LKAMKC & ONMKC & LKGr & JMKSC & MAMKKC\\ \hline
		
		BBC& \tabincell{c}{ 56.18 \\ $\pm$ 3.83 \\5.1e-28 } & \tabincell{c}{ 54.53 \\ $\pm$ 3.28 \\2.7e-34 } & \tabincell{c}{ 55.74 \\ $\pm$ 3.23 \\3.2e-32 } & \tabincell{c}{ 67.58 \\ $\pm$ 7.80 \\3.4e-08 } & \tabincell{c}{ 72.52 \\ $\pm$ 4.84 \\1.5e-04 } & \tabincell{c}{ 73.26 \\ $\pm$ 5.14 \\6.9e-04 } & \tabincell{c}{ 68.99 \\ $\pm$ 5.42 \\3.4e-11 } & \tabincell{c}{ 53.29 \\ $\pm$ 3.87 \\5.4e-33 } & \tabincell{c}{ 49.83 \\ $\pm$ 4.62 \\1.2e-32 } & \tabincell{c}{ \textbf{76.62} \\ \textbf{$\pm$ 4.649 } \\ \textbf{1.0e+00 }} \\ \hline
		RELATHE& \tabincell{c}{ 64.56 \\ $\pm$ 0.29 \\1.1e-92 } & \tabincell{c}{ 67.41 \\ $\pm$ 3.3e-14 \\0.0e+00 } & \tabincell{c}{ 59.73 \\ $\pm$ 2.96 \\7.0e-48 } & \tabincell{c}{ 56.02 \\ $\pm$ 1.07 \\2.4e-72 } & \tabincell{c}{ 63.05 \\ $\pm$ 0.03 \\6.4e-141 } & \tabincell{c}{ 71.57 \\ $\pm$ 0.14 \\1.7e-99 } & \tabincell{c}{ 63.70 \\ $\pm$ 0 \\0.0e+00 } & \tabincell{c}{ 58.28 \\ $\pm$ 0.09 \\1.0e-121 } & \tabincell{c}{ 57.18 \\ $\pm$ 0.01 \\5.7e-164 } & \tabincell{c}{ \textbf{85.35} \\ \textbf{$\pm$ 1.0e-13 } \\ \textbf{1.0e+00 }} \\ \hline
		PIE10P& \tabincell{c}{ 47.87 \\ $\pm$ 4.09 \\2.3e-27 } & \tabincell{c}{ 41.23 \\ $\pm$ 3.85 \\5.1e-31 } & \tabincell{c}{ 36.45 \\ $\pm$ 2.76 \\1.9e-34 } & \tabincell{c}{ 28.26 \\ $\pm$ 2.19 \\5.5e-39 } & \tabincell{c}{ 46.27 \\ $\pm$ 4.89 \\2.6e-26 } & \tabincell{c}{ 53.83 \\ $\pm$ 5.34 \\6.81e-21 } & \tabincell{c}{ 52.56 \\ $\pm$ 5.98 \\2.2e-23 } & \tabincell{c}{ 39.39 \\ $\pm$ 3.40 \\1.0e-30 } & \tabincell{c}{ 54.63 \\ $\pm$ 4.76 \\2.4e-20 } & \tabincell{c}{ \textbf{73.18} \\ \textbf{$\pm$ 7.46 } \\ \textbf{1.0e+00 }} \\ \hline
		COIL20& \tabincell{c}{ 62.33 \\ $\pm$ 3.44 \\3.5e-18 } & \tabincell{c}{ 61.97 \\ $\pm$ 3.80 \\9.3e-15 } & \tabincell{c}{ 60.46 \\ $\pm$ 4.06 \\2.4e-18 } & \tabincell{c}{ 61.90 \\ $\pm$ 3.44 \\8.1e-18 } & \tabincell{c}{ 58.82 \\ $\pm$ 3.78 \\7.5e-19 } & \tabincell{c}{ 58.97 \\ $\pm$ 4.43 \\3.2e-19 } & \tabincell{c}{ 59.42 \\ $\pm$ 3.230 \\5.8e-21 } & \tabincell{c}{ 57.86 \\ $\pm$ 4.34 \\7.6e-21 } & \tabincell{c}{ 63.07 \\ $\pm$ 5.83 \\9.4e-12 } & \tabincell{c}{ \textbf{72.97} \\ \textbf{$\pm$ 4.99 } \\ \textbf{1.0e+00 }} \\ \hline
		UMIST& \tabincell{c}{ 46.17 \\ $\pm$ 3.35 \\2.7e-24 } & \tabincell{c}{ 44.94 \\ $\pm$ 2.54 \\2.1e-26 } & \tabincell{c}{ 46.08 \\ $\pm$ 3.37 \\5.7e-23 } & \tabincell{c}{ 43.36 \\ $\pm$ 1.84 \\1.2e-28 } & \tabincell{c}{ 47.09 \\ $\pm$ 2.53 \\3.9e-24 } & \tabincell{c}{ 45.28 \\ $\pm$ 2.18 \\3.2e-24 } & \tabincell{c}{ 48.59 \\ $\pm$ 2.65 \\1.7e-19 } & \tabincell{c}{ 43.16 \\ $\pm$ 2.61 \\6.4e-28 } & \tabincell{c}{ 54.10 \\ $\pm$ 4.08 \\8.2e-10 } & \tabincell{c}{ \textbf{61.72} \\ \textbf{$\pm$ 5.10 } \\ \textbf{1.0e+00 }} \\ \hline
		BASEHOCK& \tabincell{c}{ 93.58 \\ $\pm$ 1.0e-13 \\0.0e+00 } & \tabincell{c}{ 94.76 \\ $\pm$ 0.025 \\7.6e-101 } & \tabincell{c}{ 94.13 \\ $\pm$ 7.8e-14 \\0.0e+00 } & \tabincell{c}{ 96.69 \\ $\pm$ 0.38 \\3.5e-17 } & \tabincell{c}{ 93.10 \\ $\pm$ 0.02 \\2.2e-111 } & \tabincell{c}{ 95.38 \\ $\pm$ 6.7e-14 \\0.0e+00 } & \tabincell{c}{ 96.74 \\ $\pm$ 7.8e-14 \\0.0e+00 } & \tabincell{c}{ 53.84 \\ $\pm$ 4.92 \\2.0e-48 } & \tabincell{c}{ 52.28 \\ $\pm$ 3.3e-14 \\0.0e+00 } & \tabincell{c}{ \textbf{97.39} \\ \textbf{$\pm$ 2.2e-14 } \\ \textbf{1.00e+00 }} \\ \hline
		Leukemia& \tabincell{c}{ 68.06 \\ $\pm$ 6.7e-14 \\0.0e+00 } & \tabincell{c}{ 60.97 \\ $\pm$ 1.38 \\8.2e-60 } & \tabincell{c}{ 62.50 \\ $\pm$ 0 \\0.0e+00 } & \tabincell{c}{ 69.11 \\ $\pm$ 4.09 \\3.3e-27 } & \tabincell{c}{ 63.89 \\ $\pm$ 3.3e-14 \\0.0e+00 } & \tabincell{c}{ 66.39 \\ $\pm$ 5.87 \\5.5e-24 } & \tabincell{c}{ 64.39 \\ $\pm$ 9.19 \\3.9e-18 } & \tabincell{c}{ 70.83 \\ $\pm$ 2.2e-14 \\0.0e+00 } & \tabincell{c}{ 67.17 \\ $\pm$ 7.09 \\1.2e-19 } & \tabincell{c}{ \textbf{81.94} \\ \textbf{$\pm$ 8.9e-14 } \\ \textbf{1.00e+00 }} \\ \hline
		ALLAML& \tabincell{c}{ \textbf{73.61} \\ \textbf{$\pm$ 1.1e-14} \\ \textbf{4.0e-01 }} & \tabincell{c}{ 66.92 \\ $\pm$ 0.87347 \\1.9e-05 } & \tabincell{c}{ 56.47 \\ $\pm$ 0.66461 \\1.2e-14 } & \tabincell{c}{ \textbf{71.83} \\ \textbf{$\pm$ 3.39} \\ \textbf{8.3e-02 }} & \tabincell{c}{ 66.67 \\ $\pm$ 3.3e-14 \\1.0e-05 } & \tabincell{c}{ 67.22 \\ $\pm$ 3.62 \\8.0e-05 } & \tabincell{c}{ 68.56 \\ $\pm$ 6.13 \\9.4e-04 } & \tabincell{c}{ \textbf{76.39} \\ \textbf{$\pm$ 2.2e-14} \\ \textbf{4.2e-01 }} & \tabincell{c}{ \textbf{71.86} \\ \textbf{$\pm$ 5.69} \\ \textbf{9.7e-02 }} & \tabincell{c}{ \textbf{75.03} \\ \textbf{$\pm$ 12.00 } \\ \textbf{1.00e+00 }} \\ \hline
		Average& \tabincell{c}{ 64.04}& \tabincell{c}{ 61.59}& \tabincell{c}{ 58.95}& \tabincell{c}{ 61.84}& \tabincell{c}{ 63.92}& \tabincell{c}{ 66.49}& \tabincell{c}{ 65.37}& \tabincell{c}{ 56.63}& \tabincell{c}{ 58.76}& \tabincell{c}{ \textbf{78.03}} \\ \hline
		
		\bottomrule
	\end{tabular}
\end{table*}

\begin{table*}
	\caption{Clustering comparison on (mean NMI)/(standard derivation)/($p$-value). The bolder results are significant better than others ($p \le 0.05$).}
	\centering \label{table:nmi-aio}\setlength{\tabcolsep}{1.0pt}
	\begin{tabular}{c c  c  c  c  c  c  c  c  c  c }
		\toprule
		Data Sets & CTSC & Coreg & RMSC & RMKKM & MKKMMR & LKAMKC & ONMKC & LKGr & JMKSC & MAMKKC\\ \hline
		
		BBC& \tabincell{c}{ 38.91 \\ $\pm$ 1.74 \\1.1e-38 } & \tabincell{c}{ 37.68 \\ $\pm$ 1.97\\2.5e-39 } & \tabincell{c}{ 37.60 \\ $\pm$ 1.67 \\1.6e-43 } & \tabincell{c}{ 50.68 \\ $\pm$ 6.07 \\3.6e-13 } & \tabincell{c}{ 45.93 \\ $\pm$ 3.47 \\4.9e-25 } & \tabincell{c}{ 50.39 \\ $\pm$ 3.56 \\1.8e-17 } & \tabincell{c}{ 47.65 \\ $\pm$ 4.57 \\7.5e-21 } & \tabincell{c}{ 38.01 \\ $\pm$ 2.35 \\1.5e-38 } & \tabincell{c}{ 24.07 \\ $\pm$ 4.93 \\2.6e-42 } & \tabincell{c}{ \textbf{59.61} \\ \textbf{$\pm$ 3.14 } \\ \textbf{1.0e+00 }} \\ \hline
		RELATHE& \tabincell{c}{ 10.31 \\ $\pm$ 0.27 \\1.3e-102 } & \tabincell{c}{ 12.72 \\ $\pm$ 5.6e-15 \\0.0e+00 } & \tabincell{c}{ 6.02 \\ $\pm$ 1.11 \\1.3e-75 } & \tabincell{c}{ 1.99 \\ $\pm$ 2.37 \\7.8e-62 } & \tabincell{c}{ 9.13 \\ $\pm$ 0.02 \\3.0e-153 } & \tabincell{c}{ 15.26 \\ $\pm$ 0.14 \\3.2e-112 } & \tabincell{c}{ 9.10 \\ $\pm$ 4.2e-15 \\0.0e+00 } & \tabincell{c}{ 3.76 \\ $\pm$ 0.11 \\8.2e-126 } & \tabincell{c}{ 2.95 \\ $\pm$ 2.1e-15 \\0.0e+00 } & \tabincell{c}{ \textbf{41.69} \\ \textbf{$\pm$ 2.2e-14 } \\ \textbf{1.0e+00 }} \\ \hline
		PIE10P& \tabincell{c}{ 55.98 \\ $\pm$ 3.19 \\7.2e-38 } & \tabincell{c}{ 48.16 \\ $\pm$ 3.58 \\8.0e-43 } & \tabincell{c}{ 40.45 \\ $\pm$ 3.14 \\5.4e-49 } & \tabincell{c}{ 30.39 \\ $\pm$ 2.67\\4.4e-56 } & \tabincell{c}{ 53.59 \\ $\pm$ 4.42 \\5.6e-36 } & \tabincell{c}{ 60.36 \\ $\pm$ 4.82 \\7.0e-29 } & \tabincell{c}{ 55.98 \\ $\pm$ 4.28 \\1.8e-34 } & \tabincell{c}{ 46.87 \\ $\pm$ 2.93 \\1.2e-45 } & \tabincell{c}{ 66.47 \\ $\pm$ 3.72 \\1.6e-28 } & \tabincell{c}{ \textbf{83.98} \\ \textbf{$\pm$ 3.76 } \\ \textbf{1.0e+00 }} \\ \hline
		COIL20& \tabincell{c}{ 75.04 \\ $\pm$ 1.53 \\1.8e-30 } & \tabincell{c}{ 75.18 \\ $\pm$ 2.07 \\2.8e-27 } & \tabincell{c}{ 74.15 \\ $\pm$ 2.09 \\9.0e-30 } & \tabincell{c}{ 73.89 \\ $\pm$ 1.87 \\2.6e-29 } & \tabincell{c}{ 74.81 \\ $\pm$ 2.41 \\9.8e-27 } & \tabincell{c}{ 71.68 \\ $\pm$ 2.25 \\3.1e-33 } & \tabincell{c}{ 72.55 \\ $\pm$ 1.69 \\7.3e-32 } & \tabincell{c}{ 69.96 \\ $\pm$ 2.26 \\3.1e-37 } & \tabincell{c}{ 78.98 \\ $\pm$ 3.39 \\1.8e-16 } & \tabincell{c}{ \textbf{86.05} \\ \textbf{$\pm$ 2.62 } \\ \textbf{1.0e+00 }} \\ \hline
		UMIST& \tabincell{c}{ 65.07 \\ $\pm$ 2.2636 \\3.0e-31 } & \tabincell{c}{ 64.80 \\ $\pm$ 1.78 \\4.8e-37 } & \tabincell{c}{ 65.35 \\ $\pm$ 2.77 \\5.2e-28 } & \tabincell{c}{ 64.13 \\ $\pm$ 1.57 \\8.0e-37 } & \tabincell{c}{ 67.23 \\ $\pm$ 1.94 \\2.9e-29 } & \tabincell{c}{ 66.63 \\ $\pm$ 2.12 \\3.3e-33 } & \tabincell{c}{ 67.36 \\ $\pm$ 1.78 \\3.3e-31 } & \tabincell{c}{ 68.61 \\ $\pm$ 1.83 \\6.8e-31 } & \tabincell{c}{ 73.09 \\ $\pm$ 1.61 \\4.9e-19 } & \tabincell{c}{ \textbf{78.91} \\ \textbf{$\pm$ 2.41 } \\ \textbf{1.0e+00 }} \\ \hline
		BASEHOCK& \tabincell{c}{ 65.60 \\ $\pm$ 1.1e-14 \\0.0e+00 } & \tabincell{c}{ 70.69 \\ $\pm$ 0.11 \\1.9e-100 } & \tabincell{c}{ 68.06 \\ $\pm$ 3.3e-14 \\0.0e+00 } & \tabincell{c}{ 80.27 \\ $\pm$ 2.03 \\2.9e-11 } & \tabincell{c}{ 63.79 \\ $\pm$ 0.09 \\1.3e-114 } & \tabincell{c}{ 73.24 \\ $\pm$ 2.2e-14 \\0.0e+00 } & \tabincell{c}{ 79.93 \\ $\pm$ 4.4e-14 \\0.0e+00 } & \tabincell{c}{ 5.11 \\ $\pm$ 3.68 \\9.2e-67 } & \tabincell{c}{ 1.72 \\ $\pm$ 0.02 \\2.3e-178 } & \tabincell{c}{ \textbf{82.73} \\ \textbf{$\pm$ 1.1e-14 } \\ \textbf{1.0e+00 }} \\ \hline
		Leukemia& \tabincell{c}{ 9.70 \\ $\pm$ 2.8e-15 \\0.0e+00 } & \tabincell{c}{ 5.02 \\ $\pm$ 1.35 \\3.0e-61 } & \tabincell{c}{ 5.86 \\ $\pm$ 1.4e-15 \\0.0e+00 } & \tabincell{c}{ 11.98 \\ $\pm$ 2.87 \\2.5e-37 } & \tabincell{c}{ 6.70 \\ $\pm$ 0 \\0.0e+00 } & \tabincell{c}{ 7.50 \\ $\pm$ 3.77 \\4.0e-37 } & \tabincell{c}{ 9.09 \\ $\pm$ 7.35 \\2.3e-22 } & \tabincell{c}{ 12.16 \\ $\pm$ 1.54e-14 \\0.0e+00 } & \tabincell{c}{ 22.65 \\ $\pm$ 4.15 \\2.2e-09 } & \tabincell{c}{ \textbf{26.94} \\ \textbf{$\pm$ 2.8e-14 } \\ \textbf{1.0e+00 }} \\ \hline
		ALLAML& \tabincell{c}{ 15.09 \\ $\pm$ 2.8e-15 \\1.3e-04 } & \tabincell{c}{ 8.72 \\ $\pm$ 0.87 \\1.7e-10 } & \tabincell{c}{ 3.65 \\ $\pm$ 0.29 \\3.8e-15 } & \tabincell{c}{ 11.89 \\ $\pm$ 3.45 \\1.7e-07 } & \tabincell{c}{ 7.80 \\ $\pm$ 3.92 \\3.2e-10 } & \tabincell{c}{ 15.70 \\ $\pm$ 8.39 \\4.6e-03 } & \tabincell{c}{ 14.39 \\ $\pm$ 1.27 \\4.7e-05 } & \tabincell{c}{ \textbf{18.63} \\ \textbf{$\pm$ 1.1e-14} \\ \textbf{5.5e-02 }} & \tabincell{c}{ 16.86 \\ $\pm$ 2.24 \\4.0e-03 } & \tabincell{c}{ \textbf{21.83} \\ \textbf{$\pm$ 11.52 } \\ \textbf{1.0e+00 }} \\ \hline
		Average& \tabincell{c}{ 41.96}& \tabincell{c}{ 40.37}& \tabincell{c}{ 37.64}& \tabincell{c}{ 40.65}& \tabincell{c}{ 41.12}& \tabincell{c}{ 45.09}& \tabincell{c}{ 44.51}& \tabincell{c}{ 32.89}& \tabincell{c}{ 35.85}& \tabincell{c}{ \textbf{60.22}} \\ \hline
		
		\bottomrule
	\end{tabular}
\end{table*}

\section{Experiments}
In this section, we conduct several experiments to evaluate the clustering performance of  our proposed algorithm on eight benchmark data sets \cite{li2010exploiting,DBLP:conf/sdm/LiDS11,zhou2015lle,gu2015efficient} from various applications, including 3 text corpora ones (BBC, RELATHE, BASEHOCK ), 3 images ones (PIE10P, UMIST, COIL20) and 2 biological ones (Leukemia, ALLAML). The detailed information of these datasets is seen in Table ~\ref{table:dataset}.
\begin{table}
	\caption{Description of the data sets}\label{data}
	\begin{tabular}{c  c  c  c}
		\toprule
		Dataset & \# instances & \# features & \# classes\\
		\midrule
		BBC&737&1000&5\\
		RELATHE&1427&4322&2\\
		BASEHOCK&1993&5862&2\\
		PIE1OP&210&2420&10\\
		UMIST&575&644&20\\
		COIL20&1440&1024&20\\
		Leukemia&72&7070&2\\
		ALLAML&72&7129&2\\
		\bottomrule
	\end{tabular}
	\label{table:dataset}
\end{table}
\subsection{Compared Algorithms}		
We compared MAMKKC with state-of-the-art multiple kernel clustering algorithms, i.e., \textbf{CTSC} \cite{ctsc}, \textbf{Coreg} \cite{coreg_nips2011}, \textbf{RMSC} \cite{rmsc_2014}, \textbf{RMKKM}, \textbf{MKKMMR} \cite{rmkkm_ijcai2015}, \textbf{LKAMKC} \cite{lkamkc_2016}, \textbf{ONMKC} \cite{onmkkm_2017}, \textbf{LKGr} \cite{lkgr}, \textbf{JMKSC} \cite{jmksc}. It should be pointed that the code for all these 9 methods are obtained from the author's website or provided by the authors. All the code of our method can also be found at \url{https://gitee.com/csliangdu/MAMKKC}, accordingly.


\begin{figure}
	\centering
	\subfloat[]{\label{1} \includegraphics[scale=.25]{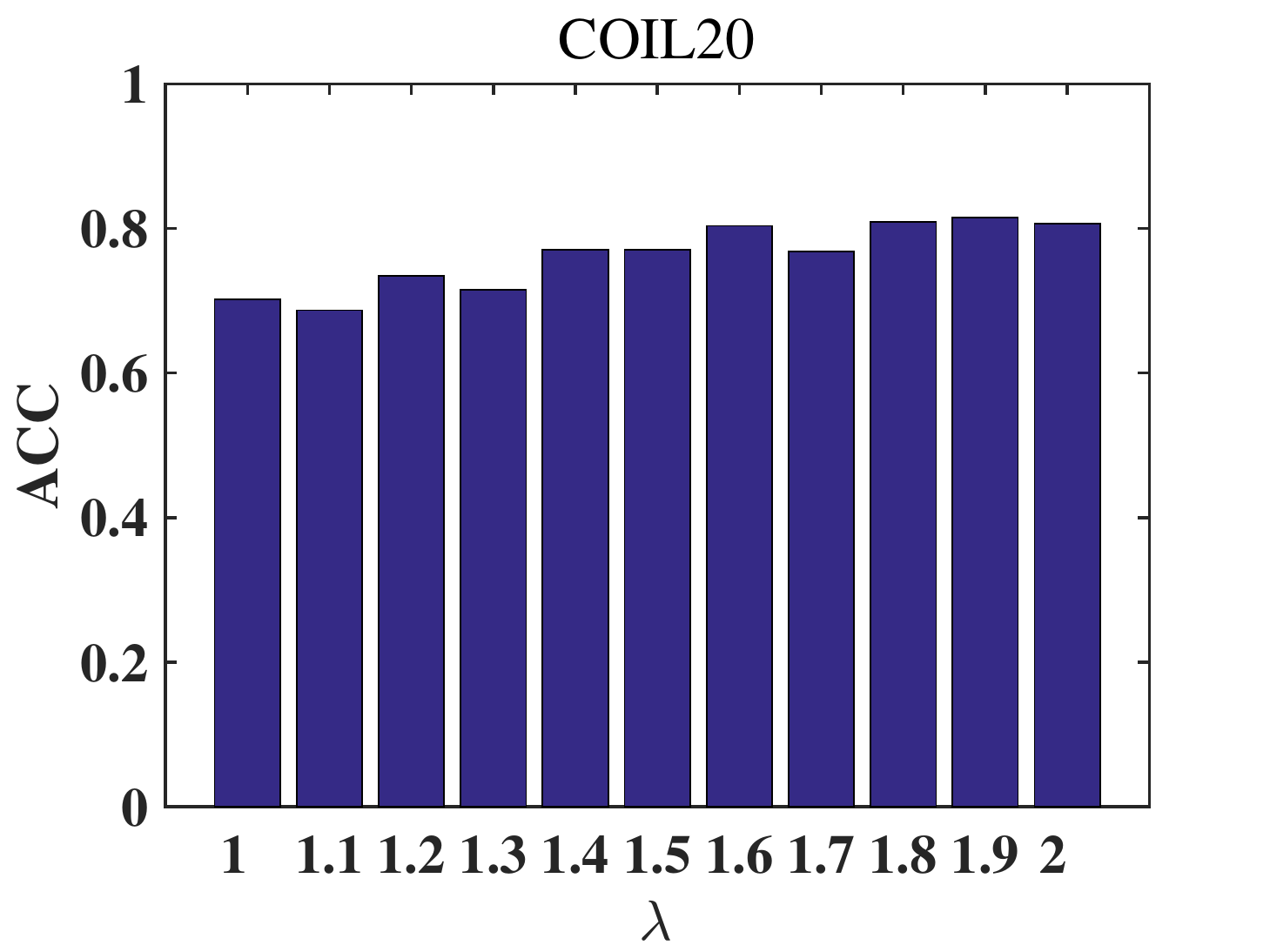}}
	\subfloat[]{\label{2} \includegraphics[scale=.25]{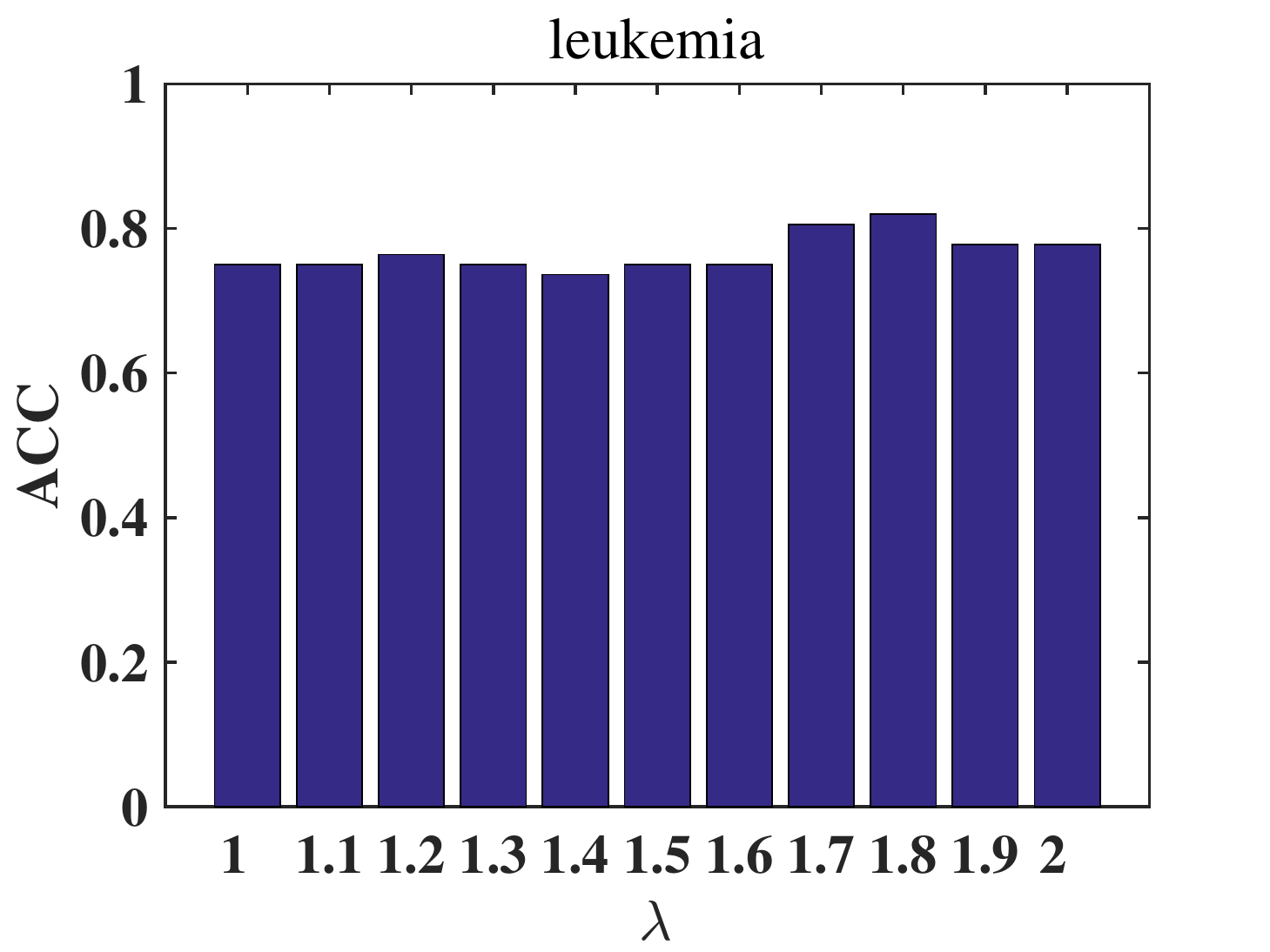}}
	
	\caption{ACC variants of $\lambda$ on COIL20 and Leukemia}
	\label{fig:sen}
\end{figure}
\begin{figure}
	\centering
	\subfloat[]{\label{fig:converge_Lung} \includegraphics[scale=.25]{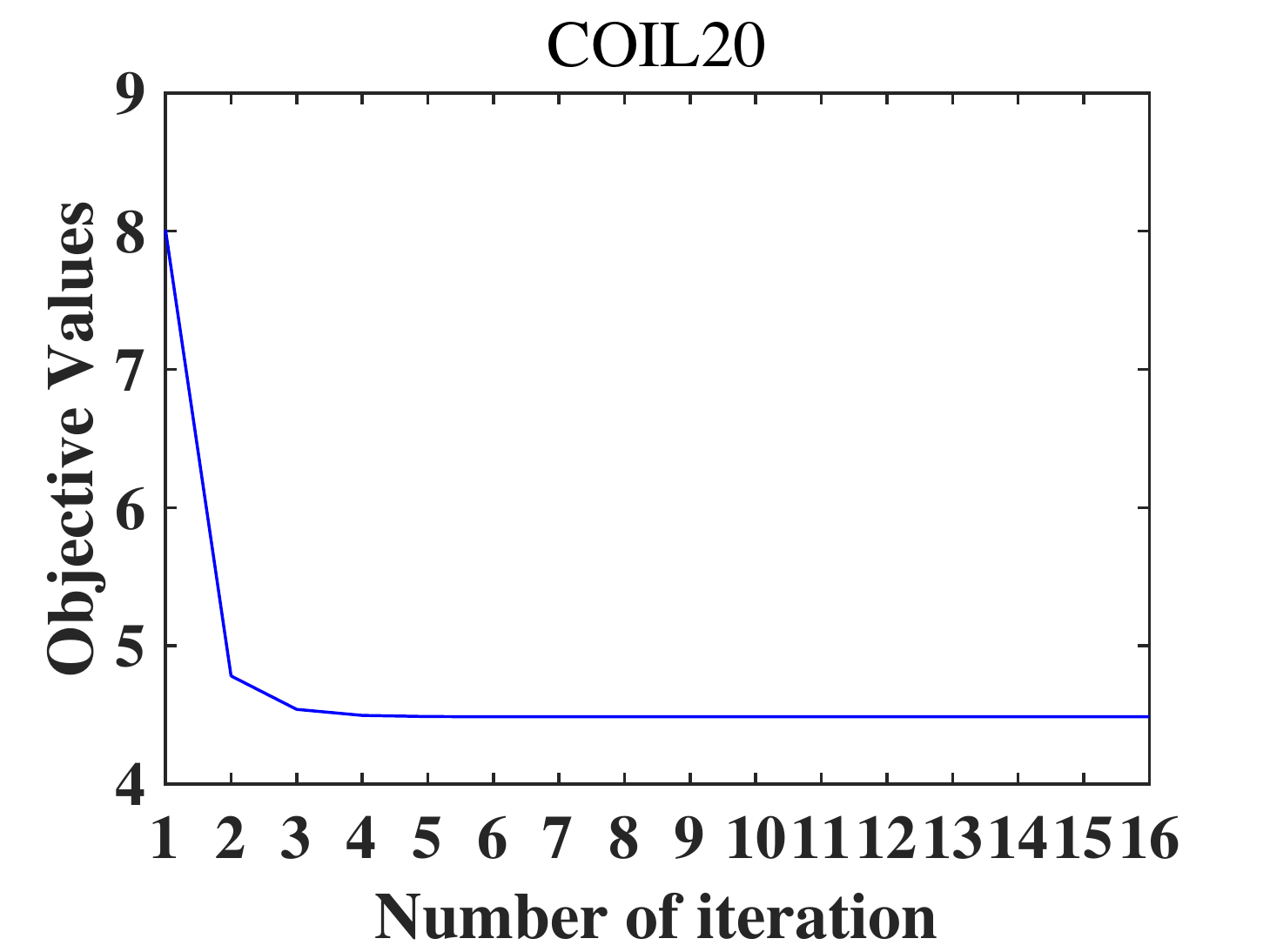}}
	\subfloat[]{\label{fig:converge_USPST} \includegraphics[scale=.25]{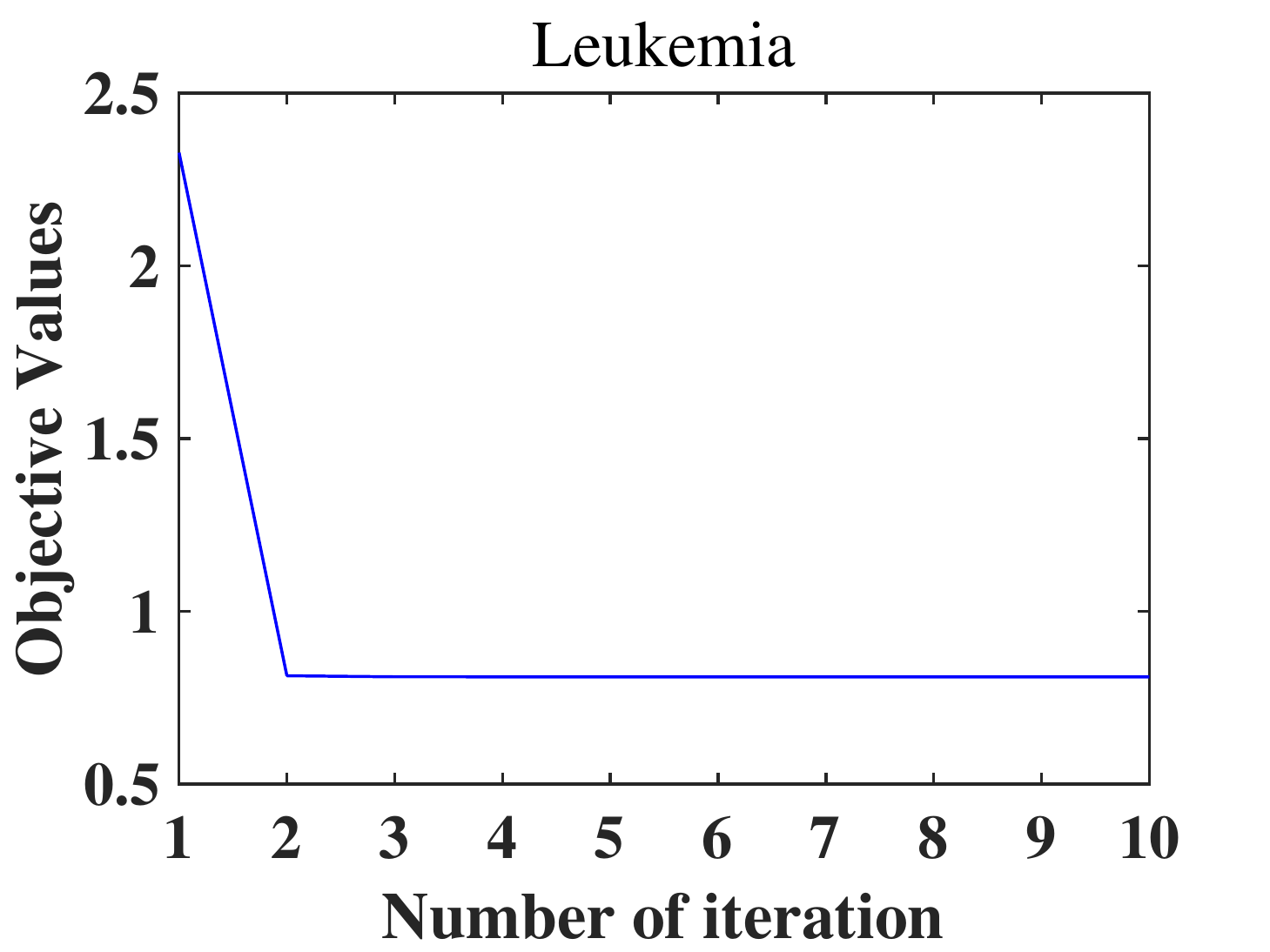}}
	\caption{The convergence on COIL20 and Leukemia.}
	\label{fig:cov}
\end{figure}
\subsection{Experimental settings}
We follow similar settings to prepare 12 base kernels according to \cite{dmkcf}. The parameters for these baselines are also set as \cite{dmkcf}. For our proposed algorithm, The parameter of proposed algorithm is chosen from  $\{1.0,1.1,\ldots\,2.0\}$. Like previous works, we also use three external measures, i.e., clustering accuracy (ACC), normalized mutual information (NMI) and Purity, to evaluate the clustering results.

\subsection{Experimental results}

For each clustering algorithm, we report the best results for each parameter corresponding to the best objective values in terms of ACC/NMI/Purity, respectively, from fifty rounds of random initialization in Table \ref{table:res-aio}. We also report the averaged results over all these 8 data sets in the last row of Table \ref{table:res-aio}. It can be seen that MAMKKC consistently outperform other state-of-the-art multiple kernel clustering algorithms. Compared with the second best averaged results, it can be seen that our method achieves $15.75\%$, $30.79\%$ and $15.92\%$ improvement in terms of ACC/NMI/Purity, respectively. These results show the effectiveness of the proposed method.

For each clustering algorithm, we also calculate the the mean ACC/NMI from fifty rounds of random initialization for each parameter and then we additionally report the best mean ACC/NMI together with the standard deviation corresponding to the optimal parameter and the $p$-value of the paired $t$-test against the best results in Table \ref{table:acc-aio}, \ref{table:nmi-aio}. Thus, each cell in Table \ref{table:acc-aio}, \ref{table:nmi-aio} include the best mean ACC/NMI, the standard deviation and the $p$-value. The best one and those having no significant difference ($p > 0.05$) from the best one are marked in bold. Again, we can observe that our method outperforms better than other MKC algorithms in most cases. And the improvements in most cases are also significant.

For all these compared multiple kernel clustering algorithms, we can observe that the ACC/NMI in Table \ref{table:res-aio} corresponding to the best objective values are generally higher than the mean ACC/NMI in Table \ref{table:acc-aio}, \ref{table:nmi-aio}.

\subsection{Parameter selection and Convergence}

For our proposed algorithm, Only one regularization parameter $\lambda$ need to be tuned. As can be seen from Figure~\ref{fig:sen}, it plot the clustering accuracy(ACC) with different values of these parameters on COIL20 and Leukemia respectively. From this figures, it can be seen   that the performance of our algorithm  is not very sensitive to $\lambda$ within relative wide ranges.

In addition, Figure~\ref{fig:cov} records the variation trend of the objective function value of our proposed method with increasing number of iterations on three data sets, i.e., COIL20 and Leukemia, respectively. As seen from these figures, the objective function is monotonically decreasing, which have also been proofed theoretically. Furthermore, the method quickly converge in less than ten iterations.

\section{Conclusions and future work}
In this paper, we proposes the multiple kernel Kmeans clustering algorithm with weighted manifold adaptive learning. The proposed MAMKKC algorithm explicitly takes into account the intrinsic manifold structure. The local geometry of the data is captured by a nearest neighbor graph. The graph Laplacian is incorporated into the manifold adaptive kernel space in which multiple kernel clustering is then performed. It can be seen that MAMKKC achieves a good performance compared to many state-of-the-art methods in the extensive experimental. 


\section{Acknowledgments}
This work is supported in part by the National Natural Science Foundation of China grant 61502289, 61806003, Shanxi Province Key R$\&$D program 201803D31199, Natural Science Foundation of Shanxi Province, China grant No.201801D221163, and Scientific and Technological Innovation Programs of Higher Education Institutions in Shanxi STIP 2016101.

\bibliographystyle{ACM-Reference-Format}
\bibliography{MAMKKC}

\end{document}